# Cultural Algorithm Toolkit for Multi-objective Rule Mining

Sujatha Srinivasan[1] and Sivakumar Ramakrishnan[2]

[1]Department of Computer Science, Cauvery College for Women, Trichy, India
`ashoksuja03@yahoo.co.in`
[2] Dept. of Computer Science, AVVM Sri Pushpam College, Poondi, Tamil Nadu, India

**ABSTRACT**

*Cultural algorithm is a kind of evolutionary algorithm inspired from societal evolution and is composed of a belief space, a population space and a protocol that enables exchange of knowledge between these sources. Knowledge created in the population space is accepted into the belief space while this collective knowledge from these sources is combined to influence the decisions of the individual agents in solving problems. Classification rules comes under descriptive knowledge discovery in data mining and are the most sought out by users since they represent highly comprehensible form of knowledge. The rules have certain properties which make them useful forms of actionable knowledge to users. The rules are evaluated using these properties namely the rule metrics. In the current study a Cultural Algorithm Toolkit for Classification Rule Mining (CAT-CRM) is proposed which allows the user to control three different set of parameters namely the evolutionary parameters, the rule parameters as well as agent parameters and hence can be used for experimenting with an evolutionary system, a rule mining system or an agent based social system. Results of experiments conducted to observe the effect of different number and type of metrics on the performance of the algorithm on bench mark data sets is reported.*

## 1. INTRODUCTION

Hybridization of techniques from various domains is an active area of research in computer science. Artificial intelligence, Swarm intelligence, and Evolutionary algorithms are called nature inspired computing (NIT) [1], while simulated annealing gets its inspiration from physical sciences. Derived from mathematics, Multi objective optimization is used in solving various optimization problems in engineering, computer science, and in particular data mining in finding optimal solutions. Whitaker [1] presents evidence suggesting that nature-inspired optimization techniques are now more frequently studied and utilized than mathematical optimization techniques and other meta-heuristics. Cultural algorithm (CA) is a class of evolutionary social system which was inspired by the evolution taking place in the society and which is used in solving optimization problems in various domains.

Classification rule mining is a class of problems where the knowledge mined is represented as "If-Then" rules and are most sought out since they are more comprehensible to the user. There are often objective and subjective measures to evaluate the rules. These measures are sometimes called the properties of the rule. The classification rules have to satisfy some of these properties to be used as a good classifier. The metrics often used for evaluating the rules are support and confidence. However there are other properties like comprehensibility and interestingness of the rule that make the classifiers more actionable to the user. But the objectives used for evaluation of rules are sometimes conflicting. For example a user may wish to have rules which are both novel and are accurate. These two objectives are conflicting since an accurate rule may not be interesting to a user and vice versa. Thus the problem of discovering rules with specific properties should be faced as a multi-objective optimization problem where the maximization or minimization of each property is one single objective.

Evolutionary algorithms (EA) are nature inspired systems and works on the strategy of survival of the fittest. Evolutionary multi objective optimization (EMOO) systems have been proposed in the literature to solve rule mining as a multi objective optimization problem. EMOO systems allow differing trade-offs to be incorporated into a multi objective problem. However research gap exists in giving the user freedom to control most of the parameters of an evolutionary multi objective system especially for classification rule mining. Therefore a Cultural Algorithm Toolkit for Classification Rule Mining (CAT-CRM) is proposed in the current study for interactive knowledge discovery. CAT-CRM combines the strengths of Evolutionary Computing, Social computing, Data mining and Artificial Intelligence in a Cultural algorithm framework. Cultural algorithm (CA) is an evolutionary algorithm that was introduced by Reynolds in 1994 [2] inspired by the social learning occurring in the society. CA best represents social systems in which agents thrive to optimize their utilities using various types of knowledge sources (KS) known as belief space. CA consists of two levels of evolution: the microevolution in a population space and the macroevolution in the belief space. The experiences of individuals in the population space are used to generate problem solving knowledge that is to be stored in the belief space which then manipulates the knowledge and in turn guides the evolution of the population space by means of an influence function [3]. Cultural algorithms have been used for modelling the evolution of complex social systems and for solving various optimization problems. The problem and related work on interactive evolutionary multi objective systems for rule mining and a short review of cultural algorithms is discussed in Section 2. Section 3 describes the proposed CAT-CRM. Experiments were conducted to see the influence of the number of metrics on the outcome of the system. The number of unique rules created by the system, the number of dominators returned by the system for difference combinations of metrics, the time taken for different number of metrics and the accuracy of the algorithm in classifying unknown data instances were observed for three different bench mark data sets. Section 4 discusses experiments and results. Section 5 concludes with future work.

## 2. PROBLEM AND RELATED WORK

### 2.1 The Problem

*Given a data source, the problem is presenting the user with a system which allows the user to control the various parameters of the system including evolutionary parameters, rule parameters and agent parameters so that the user can experiment with the system to find the influence of these parameters in discovering rule sets with differing tradeoffs in the solution and thus allowing the user to choose the best and in turn converting the knowledge discovered into actionable knowledge.*

### 2.2 Aims of the study

 i. Incorporating user preferences to control various parameters of an EMOO system for experimenting with classification rule mining as a multi objective optimization problem.

ii. Adding knowledge to Evolutionary algorithms which are blind search methods to improve the performance of the system for finding better solutions.

## 2.3 Related work

Evolutionary computing has been used extensively in data mining. Evolutionary algorithms perform a global search and are convenient for parallelization [4]. They are robust search methods that adapt to the environment and can discover interesting knowledge that will be missed by greedy algorithms [5]. Also they allow the user to interactively select interesting properties to be incorporated into the objective function providing the user with a variety of choices [6]. Thus Evolutionary algorithms are very suitable for multi-objective optimization since they allow various objectives to be simultaneously incorporated into the solution.

Participation of the user in the process of discovering knowledge is essential to improve the chance that discovered knowledge will be actually useful for the user [7]. Some systems allow the user to specify the metrics for optimization and/or the threshold values for rule selection while a very few systems allow the user to interact with the system during execution [8]. Iglesia et al. [6] [9], propose the use of multi-objective optimization evolutionary algorithms, to allow the user to interactively select a number of interest measures and deliver the best nuggets. Where in Iglesia et al. [9] propose to use Pareto-based MOEA to deliver nuggets that are in the Pareto optimal set according to some measures of interest which can be chosen by the user normally based on domain or expert knowledge. In Reynolds and Iglesia [10], the user selects a subset of the class of interest where the user is presented with a set of descriptions about the class. Presenting the user with a diverse set of rules is another area of data mining research which has been the basis of multi objective optimization in rule mining. The use of modified dominance relations have been used in [10] to increase the diversity of rules presented to the user and clustering techniques have been used in the presentation large sets of rules generated. The algorithm also considers misclassification cost and rule complexity as measures which are allowed to be controlled by the users. Whereas Reynolds and Iglesia, [11] allow the user to choose the mutation rate. In [12] the user is allowed to specify the goal attribute that is of interest to him which is used for mining highly predictive and comprehensible classification rules from large databases. Giusti et al. [13], allow the user to select a set of rules with specific properties in each generation to be used in subsequent generations. The multi objective algorithm proposed by Zhao [14] allows the decision maker to specify partial preferences on the conflicting objectives, such as false negative vs. false positive, sensitivity vs. specificity, and recall vs. precision to reduce the number of alternative solutions. This is one of a few systems which present the user with a graphical user interface. The user is allowed to choose a familiar visualization method including a ROC curve, sensitivity-specificity, precision-recall and false positive- false negative trade-offs to be visualized. The system also allows the user to visualize the progress of the evolution of solutions such that the decision maker can decide to stop the procedure when satisfactory solutions have been found or when the solutions on the front appear to have stabilized.

Apart from support, coverage and confidence of rules there are other measures that make the classifier appealing to the user such as surprisingness, interestingness, and comprehensibility of the rules. Moreover there are application specific metrics. For example sensitivity and specificity are rule metrics which are used in medical domain while precision and recall are measures used in information retrieval problems. MEPAR-miner (Multi-Expression Programming for Association Rule Mining) for rule induction is proposed in [4] which uses sensitivity and specificity of rules to define their fitness function. Reynolds et al., [15] describe the application of a multi-objective Greedy Randomized Search Procedure to rule selection, where previously generated simple rules are combined to give rule sets that minimize complexity and misclassification cost. A hybrid approach that combines a meta-heuristic and an exact operator is presented by Khabzaouil et al., [16] not only for finding non frequent rules but interesting ones also. The authors of [17], [18] propose Pitts-DNF-C, a multi-objective Pittsburgh-style Learning Classifier System that evolves a set of fuzzy rules for classification tasks. The system is explicitly designed to create consistent, complete, and compact rules for the user to comprehend.

However evolutionary algorithms so far discussed are blind search methods and allow only partial preferences of the user to be incorporated into the system. Research gap exists in incorporating knowledge and user preferences into evolutionary systems to improve the performance and usability of the systems.

**2.3.1. A short review of Cultural algorithms**

Cultural algorithm is an evolutionary algorithm which is mostly applied in solving numerical function optimization problems and which has a set of five Knowledge sources for representing various primitive knowledge's and works on the strategy of survival of the fittest. The agents in the system affect the various Knowledge sources and the KS's in turn influence the agents thus directing them towards an optimal solution. Reynolds et al., [19], use cultural algorithm to solve numerical optimization problems to study the micro and macro evolution of the individuals and the system. The individuals are provided with five types of knowledge which are said to be primitive knowledge used by most living species including human beings. Cultural algorithm has been used in rule based systems. Sternberg and Reynolds [20] use an evolutionary learning approach based on cultural algorithms to learn about the behaviour of a commercial rule-based system for fraud detection. The learned knowledge in the belief space of the cultural algorithm is then used to re-engineer the fraud detection system. Lazar and Reynolds, [21] have used genetic algorithms and rough sets for knowledge discovery. Reynolds et al., [22] use decision trees to characterize location decisions made by early inhabitants at Monte Alban, a prehistoric urban centre, and have injected these rules into a socially motivated learning system based on cultural algorithms. They have then inferred an emerging social fabric whose networks provide support for certain theories about urban site formation. Reynolds and Mostafa, [23] propose a Cultural Algorithm Toolkit which allows users to easily configure and visualize the problem solving process of a Cultural Algorithm. The proposed system is applied in solving predator/prey problem in a cones world environment and engineering design.

This paper makes a unique contribution by providing a Cultural Algorithm Toolkit for Classification Rule Mining (CAT-CRM) where the user can control three types of parameters namely the evolutionary parameters, the rule parameters and agent parameters. The system is designed considering rule mining as a multi-objective optimization problem and providing an evolutionary computation approach. The evolutionary parameters that can be controlled include the population size, the number of generations, crossover rate and mutation rate. The rule parameters that can be specified by the user are the rule metrics for optimization and a rule schema. The agent parameters include the number of agents of each type namely cautious, imitator and risk taker explained in later sections. Moreover by incorporating the various knowledge sources using a cultural algorithm framework, knowledge has been added to the otherwise blind evolutionary algorithm.

## 3. CULTURAL ALGORITHM TOOLKIT FOR CLASSIFICATION RULE MINING (CAT-CRM)

Cultural Algorithm which derives from social structures, and which incorporates evolutionary systems and agents, and uses various knowledge sources for the evolution process better suits the need for solving multi objective optimization problem and has been used in different domains.

CA has three major components: a population space, a belief space, and a protocol that describes how knowledge is exchanged between the first two components. The population space can support any population-based computational model, such as Genetic Algorithms, and Evolutionary Programming [22]. A Cultural Algorithm Toolkit for Classification Rule Mining (CAT-CRM) is proposed in the current study. Fig 1 gives the flow chart of the proposed cultural algorithm. The various components of the cultural algorithm like the belief space, the population space enabled by the evolutionary strategy and the communication protocol for knowledge exchange between the population and belief space through the influence and acceptance phase are discussed below.

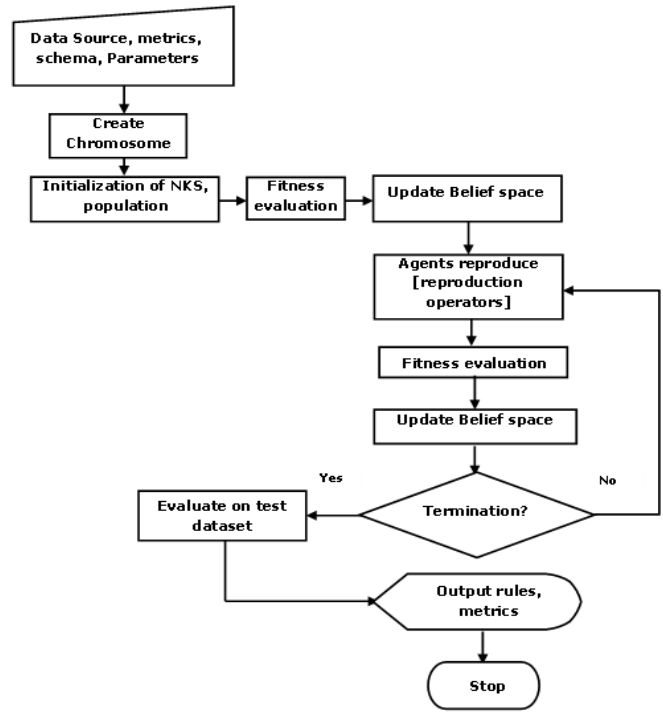

Fig 1 Flow chart of the cultural algorithm

## 3.1. The Belief space

The belief space comprises of the five knowledge sources. For the rule optimization problem the five knowledge sources are modified to hold different types of knowledge or Meta data used in solving the problem. Further an additional KS has been added to hold the rules.

*Normative KS (NKS)*: NKS contains the possible values that the attributes can take. This information is gathered from the training data set. The normative knowledge source is used to store the maximum and minimum values for numeric attributes. For nominal or discrete attributes a list of possible values that the attribute can take are stored. The normative KS is used by the agents during mutation when an attribute which is chosen to be mutated is replaced by a value taken from the NKS.

*Situational KS (SKS)*: SKS consists of the best exemplar found along the evolutionary process. It represents a leader for the other individuals to follow. Agents use these examples for crossover and mutation in subsequent generations. In the current implementation user specified schema is stored in SKS.

*Domain KS (DKS)*: DKS contains the vector of rule metrics for each rule. It is updated whenever better rules are accepted into the population at the end of each generation. The domain KS is used by the agents to choose best rules for reproduction using the Pareto optimization strategy.

*Topographical KS (TKS)*: TKS is used to store the difference or distance between two rules for the purpose of discovering diverse set of rules to avoid local optima. Hence topographical KS can be used to create novel and interesting rules by using the dissimilarity measure of an individual. This KS is updated at the end of each generation. The topographical knowledge contains a rule pair and their dissimilarity measure. Here since the algorithm takes discrete values the attribute values are compared and a value of 1 is assigned to the attributes with same values and a value of 0 is assigned to dissimilar values. The number of 1's is counted and assigned as the dissimilarity measure for the pair of rules.

*History KS (HKS)*: HKS records in a list, the best individual found at the end of each generation. Evolutionary algorithms are termed as memory less since they do not retain memory of previous generations. Cultural algorithm renders memory to the evolutionary strategy in a systematic way by using the five knowledge sources. History knowledge can be used to store elite individuals of each generation thus maintaining memory across generations. The history KS is updated at the end of each generation where the elite individuals from the rule KS are chosen according to the optimization strategy.

*The Rule KS (RKS)*: The original cultural algorithm is extended to hold another knowledge source namely the Rule KS (RKS) in order to hold the rules. The other KS's hold a pointer which is the Rule Id to the rules in the rule KS along with other values. The rule KS is added to the CA in order to render it to solve the problem of rule mining thus making CA as Extended CA or ECA. The representation of RKS is similar to that of the HKS.

*Social Agents:* The proposed ECA is also extended by adding cognitive traits to the agents which is less found in the original CA. The agent's are distinguished by assigning a cognitive trait namely risk taker or imitator or cautious. The agents use this trait in the selection of parents for reproduction using different knowledge sources.

### 3.2 Influence phase and the Acceptance phase

The *influence function* decides which knowledge sources influence individuals. In the proposed system this is left to the agents. In the proposed ECA the agents use their social trait namely risk taker or imitator or cautious to choose parents for reproduction. Risk takers use knowledge from any of the knowledge sources at random while cautious agents use only the historical knowledge source containing the elite individuals. The imitators use the SKS and RKS to create individuals which are similar to the example specified by the user. The normative knowledge source which stores the possible attribute values is used by all the agents during the mutation operation. The *acceptance function* determines which individuals and their behaviors can impact the belief space knowledge [19]. At the end of a generation (iteration), the agents return their best individuals along with a vector of rule metric values. The individuals are accepted into the belief space based on the objective vectors stored in DKS using Pareto optimality strategy. Dominators in the user specified metrics are stored in HKS while low performing individuals are removed from DKS. Thus HKS, TKS and DKS are updated at the end of each generation. The new values in these KSs then influence the population space enabling macro evolution. The process of agent's selection, reproduction, evaluation forms a generation.

## 3.3 Evolutionary strategy

Genetic algorithm (GA) is by far the most used evolutionary strategy which is also used in the current study. The various attributes of the GA used are discussed below.

### 3.3.1 Chromosome representation

The chosen data records are converted into fixed size chromosomes and represented as a vector of attribute values. The system uses high level encoding where the attribute values are used as they appear in the data source. The relational operators are not included in the genotype and thus are not involved in the reproduction. This representation avoids use of different types of reproduction operators for different parts of the chromosome. In the current study the class attribute is also included in the chromosome during the training phase. During the test phase classes are assigned to individuals as follows: If more than 75% of the values in the antecedent part are equal in the rule created and the test data instance then that class is assigned. If more than one rule covers the test instance then the maximum occurring class label that covers the rule is assigned otherwise maximum occurring class in the data set is assigned.

### 3.3.2. Population initialization

Evolutionary systems work on a population of individuals. Population initialization is an important aspect that decides the overall performance of the algorithm. Casillas et al., [17] state that the initialization procedure has to guarantee that the initial individuals cover all the input examples from the training data set. In the current study maximum and minimum chromosomes are used as seeds to create initial population. That is, the initialization procedure uses two initial chromosomes as seeds where one chromosome contains the minimum value of all the attributes and the other seed contains the maximum attribute values. These maximum and minimum seeds undergo reproduction and fill the population space.

### 3.3.3. Reproduction operators

The operators used for reproduction are selection, crossover and mutation.

*Selection strategy*

Agents use their social traits in choosing the individuals for reproduction. The agent with the social trait of risk taking chooses rules using any of the knowledge sources at random. The cautious agents choose individuals from historical KS consisting of the elite ones, while imitators use rule schema specified by the user from the situational KS. In this way, knowledge based selection is used rather than random selection. This kind of selection strategy aids in creating not only interesting knowledge but also a diverse set of solutions using the various KS's.

*Crossover*

One point crossover is used. Initially two individuals are chosen at random from the population. A crossover point which is a random integer whose value is less than the size of the chromosome is chosen at random and the contents of the chromosome after the crossover point are swapped. Crossover produces two children.

*Mutation*

Mutation operates on individual values of attributes in the chromosome. A mutation point is chosen similar to that of the crossover point which is a random integer whose value is less than the chromosome size. The value of the attribute at that point is replaced by another value depending upon the type of the value. For nominal and/or discrete attributes the value to be replaced is chosen at random from a list of available values from NKS. If the value is continuous, a random value in a specified range of minimum and maximum values so far encountered is generated and used for reproduction. A list of values for discrete and nominal attributes and lower and upper bound for real valued attributes is stored in the normative knowledge source.

### 3.3.4. Parameters

The parameters that are to be considered and greatly influence the algorithm performance are the crossover rate and the mutation rate. Also the population size and the number of generations or the termination condition are parameters of importance. Table 1 gives the summary of parameters used in the experiments.

### 3.4. Optimization strategy/Fitness evaluation

The optimization or multi objective optimization strategy forms the acceptance phase of the cultural algorithm. The ultimate objective of multi-objective algorithms is to guide the user's decision making, through the provision of a set of solutions that have differing trade-offs between the various objectives [24], and thus the user must be involved in the process of discovering rules. Therefore in the proposed system the user is allowed to control the system by specifying most of the attributes of the system including the rule metrics (objectives), the rule schema, and other parameters as discussed earlier. The user can choose any combination of metrics including coverage, support, confidence, interest, surprise, precision, recall/sensitivity, specificity and a difference measure that stores the difference between the rule and the user specified schema. Pareto optimality and ranking composition methods are the frequently used optimization strategies. In the current study Pareto optimality has been used as the optimization strategy to select elite individuals. Pareto optimality is an optimization strategy that uses comparison of the metrics represented as a vector. An individual "A" is said to be better than another individual "B" if "A" is better than "B" in all the metric values or equal to "B" in all but one metric and better at least in one metric value. This is enabled by the use of Domain KS which stores the rule metrics as fitness vectors. The entries in the DKS are compared with each other and the best performers in all the metrics are returned as dominators. The dominators form the Pareto front found in the Historical KS at the end of the algorithm execution.

Table 1 Parameters

| Parameters | Values |
|---|---|
| Crossover rate | 80% |
| Mutation rate | 20% |
| Stopping criteria | No. of generations |
| Population size | 200(Iris), 300(LJB) and 500(WBC) |
| Initialization process | Seeding |
| Optimization strategy | Pareto optimality |
| Metrics | Coverage, Confidence, Interest, Surprise and Rule difference |

## 4. EXPERIMENTS, RESULTS AND DISCUSSION

### 4.1 Experiments

Experiments were carried out to study the influence of the number of parameters on the performance of the algorithm. Three most used data sets from the UCI machine learning data bases [25], the Iris data set, the Ljubljana Breast Cancer (LJB) data set and the Wisconsin Breast Cancer (WBC) data sets were used for the experiments. Table 2 summarizes the data sets information.

Table 2 Data sets information

| Data Set | No. of data instances | No. of independent Attributes | No. of classes |
|---|---|---|---|
| Iris | 150 | 4 | 3 |
| Ljubljana BC (LJB) | 277 | 9 | 2 |
| Wisconsin BC (WBC) | 683 | 9 | 2 |

The attributes in the Iris and Ljubljana data set are continuous and categorical values. The continuous values were discretized by dividing the values into intervals and assigning a numerical integer to the intervals using simple equal width binning. The attributes information is summarized in Table 3 and Table 4 for the Iris and the LJB data set respectively. For the Wisconsin data set each of the 9 independent attributes take values ranging from 1 to 10 and were taken as is.

Table 3 Attributes, intervals and discrete values assigned to Iris data

| Attributes | Intervals & Values |
|---|---|
| Sepal length | (-, 5.5): 1, (5.6, 6.8): 2, (6.9, -):3 |
| Sepal width | (-, 2.8): 1, (2.9, 3.7): 2, (3.8, -): 3 |
| Petal length | (-, 3.0): 1, (3.1, 5.0): 2, (5.1, -): 3 |
| Petal width | (-, 0.8): 1, (0.9, 1.7): 2, (1.8, -): 3 |
| Class: Iris flower | Iris setosa : IS, Iris versicolour : IV, Iris virginica : IVG |

Table 4 Attributes, intervals and discrete values assigned to LJB data

| Attributes | Intervals & Values |
|---|---|
| Age | (<=39):1, (40-49):2, (50-59):3, (>=60):4 |
| Menopause | (Lt40):1, (Ge40):2, (Premeno):3 |
| Tumor-size | (0-9):1,(10-19):2,(20-29):3, (30-39):4, (40-49):5, (50-59):6 |
| Inv-Node | (0-2):1, (3-5):2, (6-8):3, (9-11):4, (12-14):5, (15-17):6, (24-26):7 |
| Node-Caps | 1(Yes), 0(No) |
| Deg-Malig | 1,2,3 |
| Breast | 1(Right), 0(Left) |
| Breast-Quad | 1(Left-Up), 2(Left-Low),3(Right-Up),4(Right-Low), 5(Central) |
| Irradiat | 1(Yes), 0(No) |
| Class | 1(recurrence-events), 0(no-recurrence-events) |

## 4.2 Results and Discussion

The performance of the algorithm in classifying unknown data instances was observed. Also the number of unique rules created by the algorithm (as found in RKS), the number of dominators returned by the algorithm on the various objectives (as found in HKS) and the time taken by the algorithm to produce, choose and return the dominators was observed. The parameters chosen for study are Coverage and Confidence for accuracy, Interest, Surprise as defined in [26] and Rule difference for testing the novelty of the rules. The set of metrics taken for optimization are (Coverage, Confidence), (Coverage, Confidence, Interest, Surprise) and (Coverage, Confidence, Interest, Surprise, Rule difference). The metrics are calculated as follows.

If R: A→C represents the rule. Let A be the set of all instances that satisfy the antecedent part and C be the set of all data instances that satisfy the consequent part. Let |S| be the cardinality of a set S and N the sample size. Then coverage, confidence, interest and surprise of a rule are defined as in equations (1), (2), (3) and (4) as follows:

Coverage(R)    =    |A and C|/|C|                                                                  (1)

Confidence(R)  =    |A and C|/|A|                                                                  (2)

Interest(R)    =    N*|A and C|/|A|*|C|                                                            (3)

Surprise(R) =  (|A and P| - |A and (not P)| )/|not P|                                              (4)

The rule difference is calculated by comparing the rule with the rule schema specified by the user. The number of values that differ in the rule and the rule schema are counted and returned as the rule difference. It is used to surprise the user.

The number of rules in RKS (unique rules), HKS (Dominators in the objectives) and the CPU time in milliseconds for 2, 4 and 5 objectives are averaged over ten runs of the algorithm are summarized in Table 5. Fig 2 plots the number of rules in RKS, HKS and the time against the number of objectives. Fig 3 shows sample set of dominators returned by the algorithm taking the two metrics of coverage and confidence as objectives. From Table 4 and Fig 2(a), (b) and (c) it is interesting to note that the number of unique rules created by the algorithm increases as the number of objectives increases for the Iris data set while for the Ljubljana breast cancer data set and Wisconsin breast cancer data set the number of unique rules created decreases as the number of objectives increases. This is also the case with that of the time taken by the algorithm to reach termination condition and return the rules. The time taken by the algorithm increases as the number of objectives increases for the Iris data set while the time taken decrease as the number of objectives increases for the LJB and WBC data sets. The reason for this might be that for the Iris data set since the number of attributes is less in number more number of rules cover the train data and get selected to go to the next generation. While for the LJB and WBC data sets, the number of attributes is 10, therefore the number of rules that cover the train data and qualify to go to the next generation becomes less. This is also the reason for the time taken for the algorithm to reach termination condition. For the Iris data set since more rules are chosen as candidates for the next generation the number of comparisons of the individuals in RKS with the train data instances and the number of comparison of objective vectors in DKS for choosing best individuals increases. But for the LJB and the WBC data sets since the number of individuals that qualify for the next generation decreases, the time also decreases. As for the number of dominators returned by the algorithm, for all the data sets the number of dominators decreases as the number of objectives decreases. When the number of objectives was increased to six objectives, the algorithm literally returned no rules as dominators.

Table 5 Multi-Objective optimization

| Data Sets | Data set sizes | | Average over ten runs | | | | |
|---|---|---|---|---|---|---|---|
| | No. of instances | No. of attributes | No. of objectives | No. of rules (RKS) | No of Rules (HKS) | CPU time (Milliseconds) | Accuracy % |
| Iris | 150 | 4 | 2 | 91.7 | 6.6 | 1472.5 | 95.8 |
| | | | 4 | 95.3 | 5.8 | 2023.5 | 96.8 |
| | | | 5 | 95.7 | 1.58 | 2726.7 | 92.8 |
| LJB | 277 | 9 | 2 | 134.8 | 6.6 | 7295.5 | 94.68 |
| | | | 4 | 125.2 | 2.1 | 2906.9 | 95.13 |
| | | | 5 | 115.1 | 1.4 | 2511.0 | 63.44 |
| WBC | 683 | 9 | 2 | 317.5 | 16.2 | 13154.0 | 94.87 |
| | | | 4 | 211 | 11.5 | 11710.0 | 95.18 |
| | | | 5 | 214.6 | 3.1 | 10285.1 | 93.55 |

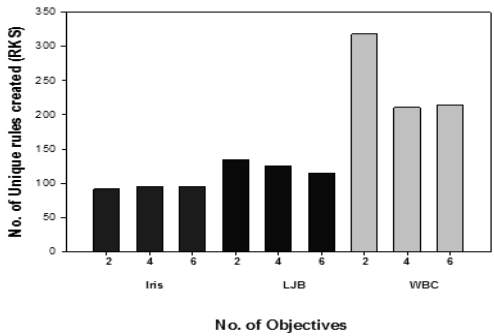
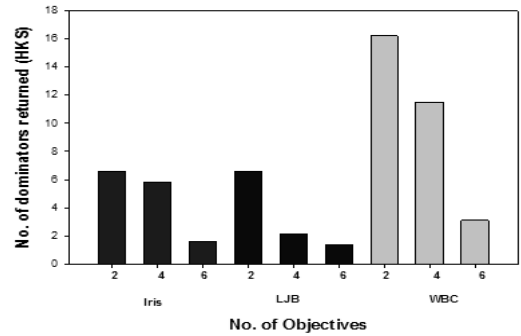

2(a) No. of Objectives VS RKS            2(b) No. of Objectives VS HKS

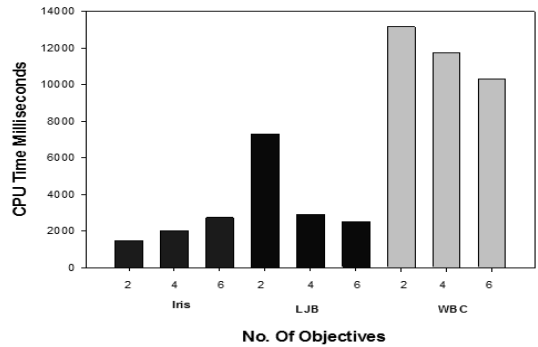

2(c) No. of Objectives VS Time

Fig 2 Number of objectives versus the number of unique rules created, number of dominators (HKS) returned and time taken in milliseconds for the three data sets.

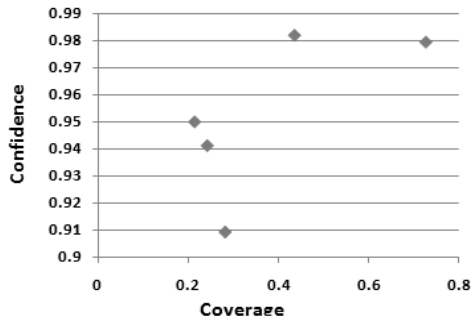 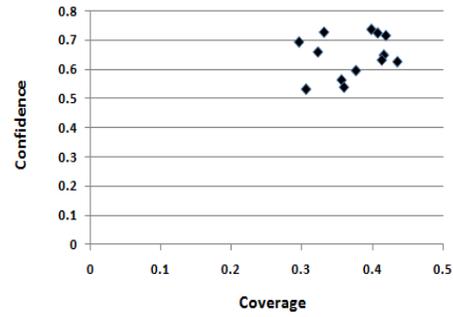

Fig 3(a) Iris  Fig 3(b) Ljubljana BC

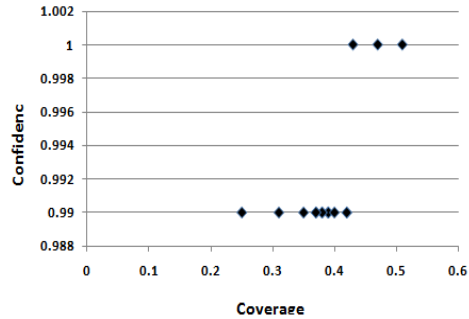

Fig 3(c) Wisconsin BC

Fig 3 Dominators in Coverage and Confidence for the three data sets

The accuracy of the rules returned on the test data set gives an interesting observation that when taking Coverage, Confidence, Interest and Surprise as objectives, the accuracy obtained is the maximum for all the three data sets. This sheds light in the direction that adding interest and surprise as additional metrics produces a compact set of rules with high accuracy. This is in accordance with the observations of Khabzaoui et al., [26] where they argue that the inclusion of the metrics of interest and surprise enables the algorithm to find rules for the data instances in the region of weaker support. It can therefore be concluded that the two measures interest and surprise can be included for finding a compact set of accurate rules. However addition of the rule difference as the fifth metric reduces the number of dominators and thus the accuracy on test data classification. Fig 3(a), (b) and (c) shows the sample set of dominators returned by the algorithm in terms of coverage and confidence. It can be observed that the algorithm is able to return good rules with high coverage and confidence. The low coverage values correspond to the rules for the classes with less number of instances. Fig 4 shows the Graphical User Interface of the Cultural Algorithm Toolkit for Classification Rule Mining (CAT-CRM), which enables the user to control various parameters of the algorithm namely, the agent parameters, evolutionary parameters and the rule parameters. The If-Then rules are displayed in the results window along with the corresponding objective vector of metric values.

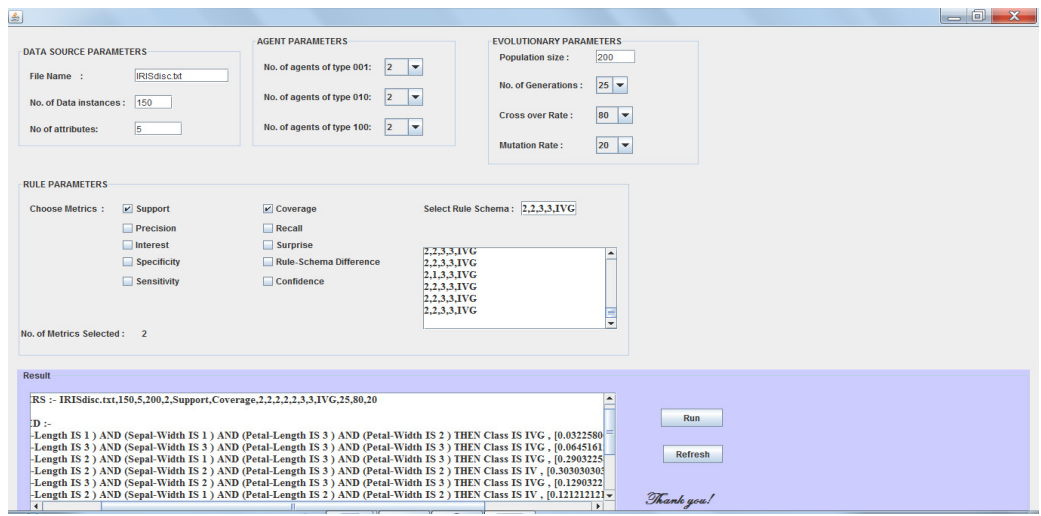

Fig 4 The Graphical User Interface of the Cultural Algorithm Toolkit for Classification Rule Mining (CAT-CRM),

## 5. CONCLUSION

In the present study a cultural algorithm toolkit is proposed for multi objective optimization of classification rules, where mining rules with specific properties is taken as a multi objective optimization problem. The tool provides a GUI through which the user can input values for different parameters. The proposed system allows the user to control rule parameters, evolutionary parameters as well as agent parameters so that the user can study the influence of various parameters on the outcome. Solutions with differing trade-offs is presented to the user from which the user can choose the best. Also the outcome of the system is a set of optimized rules which are tangible and thus can be used to evaluate the evolutionary and agent based components of the system in an efficient way. The CA enables incorporating knowledge in a systematic and principled manner into evolutionary algorithms which are blind search methods. Also incorporation of intelligent agents with cognitive traits has enabled integration of intelligent agent technology with data mining with the use of cultural algorithm, so that the system can also be used as a social system to study the dynamics of an organization or any real world social system.

## REFERENCES

[1] Whitacre J M (2011) Recent trends indicate rapid growth of nature-inspired optimization in academia and industry. Computing, 93:121–133, DOI: 10.1007/s00607-011-0154-z
[2] Reynolds, R G (1994) An introduction to cultural algorithms. In: Proceedings of the 3rd Annual Conference on Evolutionary Programming, World Scientific: River Edge, NJ, 131–139.
[3] Kendall Graham and Su Yan, (2007) Imperfect Evolutionary Systems. IEEE Transactions on Evolutionary Computation, 11(3): 294-307
[4] Baykasoglu A, Ozbakir L (2007) MEPAR-miner: multi-expression programming for classification rule mining. Eur J Oper Res 183:767–784.
[5] Freitas Alex A., (2007) A Review of Evolutionary Algorithms for Data Mining. Soft Computing for Knowledge Discovery and Data Mining, 79-111

[6] De la Iglesia B, Philpott M S, Bagnall A J, Rayward-Smith V J (2003) Data mining rules using multi-objective evolutionary algorithms. In: Proceedings of 2003 IEEE congress on evolutionary computation, 1552–1559.

[7] Freitas AA (2004) A critical review of multi-objective optimization in data mining: a position paper. SIGKDD Explor 6(2):77–86

[8] Srinivasan Sujatha and Sivakumar Ramakrishnan, Evolutionary multi objective optimization for rule mining: a review. Artificial Intelligence Review, 36(3): 205-248, DOI: 10.1007/s10462-011-9212-3.

[9] De la Iglesia B, Reynolds Alan, Rayward-Smith Vic J (2005) Developments on a multi-objective meta-heuristic (MOMH) algorithm for finding interesting sets of classification rules. In: Proceedings of third international conference on evolutionary multi-criterion optimization, EMO2005, LNCS 3410, Springer, Berlin, 826–840

[10] Reynolds A. P. and de la Iglesia, B., (2006) Rule induction using multi-objective meta-heuristic: Encouraging rule diversity. In: Proceedings of IJCNN 2006, 6375-6382.

[11] Reynolds A. P. and de la Iglesia B., (2009) A Multi-Objective GRASP for Partial Classification. Soft Computing, 13(3) :227-243.

[12] Dehuri S. and Mall R., (2006) Predictive and comprehensible rule discovery using a multi-objective genetic algorithm. Knowledge-Based Systems, 19:413–421.

[13] Giusti Rafael, Gustavo E A, Batista P A, Prati Ronaldo Cristiano, (2008) Evaluating Ranking Composition Methods for Multi-Objective Optimization of Knowledge Rules, In: Proceedings of Eighth International Conference on Hybrid Intelligent Systems, 537-542.

[14] Zhao H (2007) A multi-objective genetic programming approach to developing Pareto optimal decision trees. Decis Supp Syst 43:809–826

[15] Reynolds A P, Corne David W, De la Iglesia B (2009) A multi-objective grasp for rule selection. In: Proceedings of the 11th annual conference on genetic and evolutionary computation, GECCO'09, Montréal Québec, Canada, pp 643–650

[16] Khabzaoui M, Dhaenens C, Talbi EG, (2008) Combining evolutionary algorithms and exact approaches for multi-objective knowledge discovery. RAIRO Operations Research, 42:69–83.

[17] Casillas J, Orriols-Puig A, Bernad-o-Mansilla E (2008) Toward evolving consistent, complete, and compact fuzzy rule sets for classification problems. In: Proceedings of 3rd international workshop on genetic and evolving fuzzy systems, Witten-Bommerholz, Germany, 89–94

[18] Casillas J, Pedro Martinez AE, Benitez Alicia D (2009) Learning consistent, complete and compact sets of fuzzy rules in conjunctive normal form for regression problems. Soft Comput 13:419–465.

[19] Reynolds R G, Bin Peng, and Mostafa Ali, (2007) The Role of Culture in the Emergence of Decision-Making Roles, An Example using Cultural Algorithms. Complexity, Wiley Periodicals, Inc., 13(3): 27-42.

[20] Sternberg M and Reynolds R G, (1997) Using cultural algorithms to support re-engineering of rule-based expert systems in dynamic environments: A case study in fraud detection. IEEE Trans. Evol. Comput., 1(4):225–243.

[21] Lazar Alina and Reynolds R.G., (2002) Heuristic, Heuristics and Optimization for Knowledge Discovery. Vol., 2 (ed. Ruhul A. Sarker, Hussein A. Abbass, and Charles S. Newton) by Idea Group Publishing, USA.

[22] Reynolds, R.G., Mostafa Ali, and Thaer Jayyousi, (2008) Mining the Social Fabric of Archaic Urban Centers with Cultural Algorithms. IEEE Computer, 64-72.

[23] Reynolds, R.G. and Mostafa Z. Ali, (2007), Exploring Knowledge and Population Swarms via an Agent-Based Cultural Algorithms Simulation Toolkit (CAT). IEEE congress on evolutionary computing (CEC 2007), pp.2711-2718

[24] Reynolds A. P. and de la Iglesia B., (2007) Rule Induction for Classification Using Multi-Objective Genetic Programming. In: proceedings of 4th Int'l. Conf. on Evolutionary Multi-Criterion Optimization. LNCS 4403, 516-530.

[25] Newman D., Hettich S., Blake C. and Merz C., (1998 ) UCI Repository of Machine Learning Databases. Dept. of Information and Computer Science, Univ. of California at Irvine, http://www.ics.%20uci.edu/~mlearn/MLRepository.html.

[26] Khabzaoui M, Dhaenens C, Talbi EG (2008). Combining evolutionary algorithms & exact approaches for multi-objective knowledge discovery. RAIRO Oper Res 42: 69–83. doi:10.1051/ro:2008004

**Authors**

**Sivakumar Ramakrishnan** is Reader in the Research Department of Computer Science in AVVM Sri Pushpam College, Tamil Nadu, India since 1987. His research interests include Data mining, Human Computer Interaction and Bio-informatics. He has published a number of papers in National and International Journals. He received his PhD in Computer Science from Barathidasan University, India in the year 2005.

**Sujatha Srinivasan** received her Master's degree in Mathematics in 1993 and Master's degree in Computer Applications in 2000. She received her Master of Philosophy in Computer Science in 2004. She is Assistant Professor in the PG and Research department of Computer Science in Cauvery College for women, Tamil Nadu, India for the past ten years. She is a Research scholar in AVVM Sri Pushpam College, India. Her research interests include Simulation modeling, Data mining, Human Computer Interaction and Evolutionary computing. She has published papers in International Journals and presented papers in International Conferences.